# Towards WARSHIP:

# Combining Components of Brain-Inspired Computing of RSH for Image Super Resolution

Wendi Xu[1,3], Ming Zhang[1,2]

[1]Xinjiang Astronomical Observatories,Chinese Academy of Sciences, Urumqi 830011, China
[2]Key Laboratory for Radio Astronomy, Chinese Academy of Sciences, Nanjing 210008, China
[3]University of Chinese Academy of Sciences, Beijing 100876, China
xuwendi@xao.ac.cn, zhang.ming@xao.ac.cn

**Abstract:** Evolution of deep learning shows that some algorithmic tricks are more durable, while others are not. To the best of our knowledge, we firstly summarize 5 more durable and complete deep learning components for vision, that is, WARSHIP. Moreover, we give a biological overview of WARSHIP, emphasizing brain-inspired computing of WARSHIP. As a step towards WARSHIP, our case study of image super resolution combines 3 components of RSH to deploy a CNN model of WARSHIP-XZNet, which performs a happy medium between speed and performance.

## 1 Introduction

### 1.1 A more durable component within the evolution of deep learning: residual learning

T. Poggio observes, analyzes and predicts the evolution of deep learning from both mathematical and biological sides(which is the focus in our article) in [1]"Deep learning: mathematics and neuroscience".

He mentions that, "it is telling that several of the algorithmic tricks that were touted as breakthroughs just a couple of years ago are now regarded as unnecessary", while "some of the other ideas" such as residual learning "are more fundamental" "and likely to be more durable, though their exact form is bound to change somewhat".

In a word, he predicts that residual learning is a more durable component within the evolution of deep learning.

### 1.2 Five more durable components : WARSH

In this article, we also conjecture/predict 5 more durable components ---"WARSH", that is, (1) W: Wavelet-like transformation, (2) A: Attention models, (3) R: Residual learning(already highlighted by Poggio), (4) S: Sparsity, (5) H: preference of Hierarchical /deep models to shallow (single layer) ones.

We tend to believe that WARSH perform common, fundamental and key properties, making themselves more durable.

For example, (5) H has firm foundation mathematically [3](hierarchically local function), physically [17] and biologically[1], serving as maybe the 1st rule to be applied in deep learning.

### 1.3 WARSHIP: restrict WARSH for vision(IP), durable and complete

**Vision(IP)**: As we know, vision tasks corresponds to human visual perceptive ability, speech tasks imitate human audio perceptive ability, whereas natural language processing(NLP) performs as human cognitive ability. Deep learning based models for vision and speech via WARSH can be called intelligent perception(IP). But in this paper, partly for simplicity, suffix of IP to WARSH imposes the unique restriction that we are neither in the domain of speech(even though it can also be called IP), nor in NLP, but always in the realm of vision(IP).

**Complete**: What we should highlight is that we conjecture that WARSH are not only more durable, but also already enough and/or complete to address major/whole tasks for vision(IP), which constitute one of many significant reasons for our proposal of WARSHIP.

**WARSHIP**: Thus, we propose the concept of "WARSHIP" for vision, which is short for (1) Wavelet-like (2) Attentive (3) Residual (4) Sparse (5) Hierarchical Intelligent Perception.

### 1.4 Biological lens into WARSHIP: brain-inspired computing of WARSHIP

Science of deep learning lays at the intersect of computer science, mathematical science[3], physical science[17] and biological science[1]. Therefore, WARSHIP within deep learning are also constituted by those four parts.

To our surprise and excitement, all 5 durable and complete components in WARSHIP do have biological root respectively.

So, in this article, we view WARSHIP through biological lens, more specifically, discuss brain-inspired computing of WARSHIP.

### 1.5 Towards WARSHIP: case study of image super-resolution

Constrained in 5 durable deep learning components, restricted in machine vision tasks and inclined to biological part, our exploration towards WARSHIP starts from an interesting and nice task---image super resolution for its simplicity and fundamentality.

Image super resolution is the problem of recovering the fine details of an image ---the high end of its spectrum---from coarse scale information only---from samples at the low end of the spectrum [16]. As a subset of image super resolution, single image super resolution (SISR, which is the topic of this paper and is interchangeable with image super resolution from now on.) is the process of inferring a High-Resolution (HR) version of a Low-Resolution (LR) input image.

## 2 Main contributions

1. Durable and Complete Framework : WARSHIP.

To the best of our knowledge, we firstly summarize 5 more durable and complete deep learning components for vision, which are represented as WARSHIP, namely, (1) Wavelet-like (2) Attentive (3) Residual (4) Sparse (5) Hierarchical Intelligent Perception.

2. Brain-Inspired Computing of WARSHIP

We highlight the deep root of WARSHIP in brain science, namely, 5 key components of brain-inspired computing of WARSHIP, which also strongly proves the duration and completeness of WARSHIP in the evolution of deep learning for vision.

3. Empirical Success: WARSHIP-XZNet.

Speed-performance balance. Following 3 brain-inspired computing components in WARSHIP, namely RSH, we deploy model of WARSHIP-XZNet for image super resolution. WARSHIP-XZNet performs a happy medium between speed and performance for image super resolution.

4.Towards WARSHIP, Towards Brain Inspired Intelligence.

WARSHIP-XZNet proves to be successful combination of brain-inspired computing of RSH , making a step towards WARSHIP, towards the prospective and active direction of brain-inspired intelligence.

## 3 Biological overview of brain-inspired computing of WARSHIP

Deep learning has benefited greatly and will benefit much from neuroscience of brain. In China Brain Project [18], brain-inspired computing in deep learning can be viewed from structure, behavior, and mechanism perspectives. We follow the categorization to analyze biological overview of brain-inspired computing of WARSHIP, to strongly confirm the duration and completeness of WARSHIP during the evolution of deep learning based vision.

### 3.1 From the structure perspective：R and H

Residual learning, with 3.57% top-5 error unarguably beats human performance on the ImageNet test. Q. Liao and T. Poggio proved its root in visual cortex that ultra-deep residual network actually is "carried out" in its equivalent formulation--unrolling in time the recurrent computations in 6 visual layers [2].

Areas in visual cortex comprise six different "hierarchical biological layers" with lateral and feedback connection, which inspired the successful design of "hierarchical artificial layers" in deep learning [3].

### 3.2 From the behavior perspective：A

Attention model [4] is a concept rooted in human visual attention, which is widely used for image recognition-like, and scarcely used for image reconstruction. We conjecture that its future development for the latter will be very prospective.

### 3.3 From the mechanism perspective：W and S

Wavelet-like transformation is long believed to exist in human vision [5], with its nice properties such as sparsity and multi-scale/multi-resolution. The latter property could potentially address the problem of multi-scale object detection etc. However, mature deployments of the mechanism simulating wavelet representation in visual cortex are still on the road [6].

Sparse, a concept of interest in statistics, linear algebra, signal processing, and also in computational neuroscience and machine learning, is very closely related to the robust optimization in human and machine vision. In observation of human vision, "studies on brain energy expense suggest that neurons encode information in a sparse and distributed way, estimating the percentage of neurons active at the same time to be between 1% and 4%" [7]. And in machine vision, many designs have been inspired by sparsity in visual cortex including (1) dropout, (2) sparse rectifier and (3) filter size of 1x1 in convolutional layer with unbalanced input and output features and (4) weight decay.

## 4 A model for image super resolution: WARSHIP-XZNet

### 4.1 Specified architecture to realize WARSHIP: convolutional neural network

Theoretical framework of WARSHIP can be realized in different specified architecture of neural network.

Which architecture should we choose? The family of neural networks include auto-encoder (AE), recurrent neural network (RNN), general adversarial network (GAN), convolutional neural network (CNN) etc. We abandon GAN for its less maturation. CNNs are more well studied and are widely used not only for vision, but also for audio, natural language processing (NLP) [8].

Moreover, which building blocks in CNN should we choose? CNN is made up of layers of several types, including (1) convolutional layer, (2) activation layer, (3) pooling layer, (4) fully connected layer, (5) normalization layer, (6) de-convolutional layer. Discarding pooling layers has been proved to be important in training good generative models for image reconstruction. So only (1) convolutional layer and (2) activation layer are chosen in our model of WARSHIP-XZNet.

### 4.2 Specified mechanism behind WARSHIP-XZNet: sparse coding

Roughly deploying the deep model blind to the knowledge of the task is obviously inefficient.

VDSR [12] is blind to any partition/specification of the framework, that is, purely free without constrain/knowledge, and DCSCN [13] takes the encoder-coder formulation. FSRCNN [10] and DRCN [11] borrow the analytically fully understood sparse coding framework which was firstly introduced for SRCNN [9].

In our WARSHIP-XZNet, we also follow specified mechanism of sparse coding.

### 4.3 Implementation details of WARSHIP-XZNet

We construct 3 sub-nets, embedding sub-net (Enet), embedding image to features, inference sub-net (Inet), inferring high-resolution features from low-resolution ones, and reconstruction sub-net (Rnet), mapping back to high-resolution image. (Notation: "Conv layer" mentioned below means a convolutional layer and its following activation layer.)

In Enet, the 1st Conv layer stores the input/image, then the 2nd Conv layer performs transformation from image space to feature space, finally the 3rd shrinking Conv layer with filter size of 1x1 works as a linear combination of original features and gains a sparsity of 50%.

In Inet, our residual block with 1 convolutional layer and 1 activation layer is recurred 8 times as a special formulation of residual network. Every intermediate output of features by each recurrence will be extracted to Rnet. The parameter space of Inet is further "equivalently" enlarged by a weight layer in Rnet.

In Rnet, expanding Conv layer with filter size of 1x1 inverses the sparsity in Inet, then another Conv layer converts feature dictionary to several images, lastly a weighed Conv layer weights the images into a single image.

Our model is fully CNN of 3 + 8 + 3 = 14 layers with ReLU as activation layers. In each Conv layer, filter size is 3x3 except that in shrinking and expanding layers with 1x1. The Figure1 is the overall architecture of WARSHIP-XZNet.

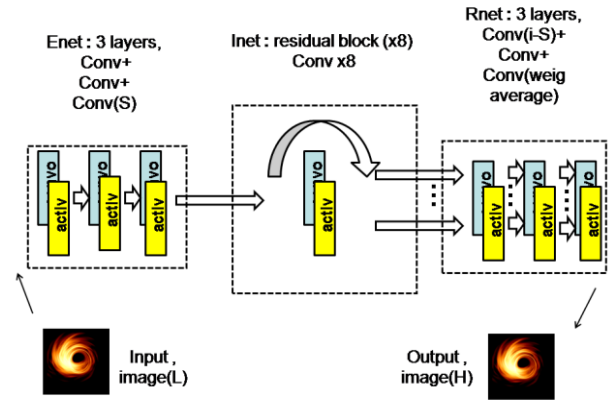

**Figure 1** overall architecture of WARSHIP-XZNet

### 4.4 Training data

We use 91 images proposed in [14] and augmented data of ScSR to train WARSHIP-XZNet.

### 4.5 Training setup

For intermediate outputs, the cost function is

$$L_1(\theta) = \sum_{i=1}^{N} \sum_{r=1}^{R} \frac{1}{2RN} \| y^{(i)} - y_r^{(i)} \| \quad (1)$$

where θ is the overall parameter set of the whole model, N is number of training samples; R is times of recurrence; $y^{(i)}$ is the i-th training sample, $y_r^{(i)}$ is the output image extracting from the r-th recurrence. In WARSHIP-XZNet, we set R to 8.

The cost function of final output is

$$L_2(\theta) = \sum_{i=1}^{N} \frac{1}{2N} \| y^{(i)} - \sum_{r=1}^{R} w_r y_r^{(i)} \|$$

where N, R, $y^{(i)}$ and $y_r^{(i)}$ is the same as equation(1), $w_r$ denotes the weight that will average images extracting from 8 recurrences into the final high resolution image.

Additional penalty of Tikhonov regularization is not equally applied on every layer:

$$L_3(\theta) = \beta \| \theta \| - \beta \| \theta(R)^{(1)} \| - \beta \| \theta(E)^{(1)} \|$$

where $\theta(R)^{(1)}$ is the parameter space of 1st layer in Rnet., and $\theta(E)^{(1)}$ is the parameter space of 1st layer in Enet.

Thus we get the overall cost function

$$L(\theta) = \alpha L_1(\theta) + (1 - \alpha) L_2(\theta) + L_3(\theta)$$

where α is the weight of $L_1(\theta)$.

Because our model is very closely to both FSRCNN and DRCN and more similar to the latter, we adopt the same training setting of DRCN.

Split training images into 41 by 41 patches with stride 21 and 64 patches, which are used as mini-batch for stochastic gradient descent. We use the method described in [15] to initialize weights in non-recursive layers, and set all weights to zero except self-connections (connection to the same neuron in the next layer) for recursive convolutions. Bias are set to zero. We choose the beginning value of learning rate to be 0.01 and then decreased it by a factor of 10 if the validation error dose not decrease for 5 epochs. If learning rate is less than $10^{-6}$, then procedure is stopped. Training roughly takes 6 days on an Many Integrated Cores (MIC) machine in Xinjiang Astronomical Observatories, Chinese Academy of Sciences.

### 4.6 Comparison with state-of-the-art algorithm: fast and accurate

We compare our model with traditional method Bicubic and deep learning based algorithms like SRCNN and DRCN. The comparison is illustrated in the PSNR table, where PSNR denotes peak signal noise ratio. Set5 is chosen as test data at up-scaling of 2.

**Table Ⅰ** Table of PSNR.

| | Traditional VS Deep learning based methods | | | |
|---|---|---|---|---|
| | Traditi. | Deep learning based methods | | |
| | Bicubic | SRCNN | Ours | DRCN |
| PSNR | 33.66 | 36.33 | 36.86 | 37.63 |

Our WARSHIP-XZNet outperforms Bicubic and SRCNN with better PSNR. Our model takes the sparsity design to gain 96 features which is quite less than 256 features in DRCN, thus ease the computation burden to speed up at the cost of some acceptable loss of PSNR comparing with DRCN.

Our fast and accurate model is especially suitable for some power-constrained applications when both memory and computational resource are limited.

## 5 Towards WARSHIP: Combining brain inspired computing of RSH

### 5.1 Highly effective S

Further details of four sparse designs mentioned in **3.3** are (1) dropout invented in the fully-connected layers of AlexNet, which helps the network to escape from substantial over-fitting; (2) sparse rectifier in the activation layer, for example, ReLU [7] and (3) filter size of 1x1 in convolutional layer [10] with unbalanced input and output features and (4)weight decay. (1), (2) and (3) are specified designs for specified layers. (4) can be applied in both convolutional layer and fully connected layer. A further exploration of (4) can be seen in our paper [19] "Theory of generative deep learning Ⅱ: probe landscape of empirical error via norm based capacity control",

(1) is suitable for image recognition--like and not for image reconstruction including SISR, so our model rules it out. Only (2), (3) and (4) are implemented in WARSHIP-XZNet. Two layers with filter size 1x1 are served as a sparsifier in Enet and an inverse-sparsifier in Rnet to undertake the sparse property of brain inspired computing. Both FSRCNN and WARSHIP-XZNet share (3) to sparsify the features for inferring in Inet.

### 5.2 Highly effective R and H

R and H are mainly expressed in the Inet.

Compared to SRCNN with only 1 layer working as Inet, WARSHIP-XZNet follow the fundamental rule of deep/hierarchical principle.

Furthermore, the formulation of hierarchical deployment are specified/constrained to be residual learning with recurrent shared residual blocks.

Residual networks with weigh sharing like our WARSHIP-XZNet corresponds to homogeneous, time-invariant systems, which is biologically-plausible, with orders of magnitude less parameters, at the same time retain most of the performance of the corresponding deep residual network [2].

## 6 Conclusion and future works

Theoretically, within deep learning framework, to the best of our knowledge, we firstly summarize 5 more durable and complete deep learning components for machine vision, which are represented in WARSHIP, namely, (1) Wavelet-like (2) Attentive (3) Residual (4) Sparse (5) Hierarchical Intelligent Perception.

We also present a biological overview of WARSHIP, focus on its side of brain-inspired computing.

Empirically, following 3 components in WARSHIP, namely RSH(Residual, Sparse and Hierarchical), we deploy model of WARSHIP-XZNet for image super resolution. WARSHIP-XZNet performs a happy medium between speed and performance for image super resolution

Our WARSHIP-XZNet proves to be successful combination of 3 components of brain-inspired computing, thus making a step towards WARSHIP, towards the prospective and active direction of brain-inspired intelligence.

In the future, we should develop more efficient designs of W and A , namely, wavelet-like and attentive design armed with inspiration from brain science.

### Acknowledgements

This work was supported by the National Basic Research Program of China (2012CB821804 and 2015CB857100), the National Science Foundation of China (11103055 and 11773062) and the West Light Foundation of Chinese Academy of Sciences (RCPY201105, 2017-XBQNXZ-A-008 and 2016-QNXZ-B-25).